\documentclass[11pt]{article}

\usepackage[final]{acl}

\usepackage{times}
\usepackage{latexsym}

\usepackage[T1]{fontenc}

\usepackage[utf8]{inputenc}

\usepackage{microtype}

\usepackage{inconsolata}

\usepackage{graphicx}

\usepackage{microtype}
\usepackage{hyperref}
\usepackage{url}
\usepackage{booktabs}
\usepackage{algorithm}
\usepackage{algpseudocode}
\usepackage{amsmath}

\usepackage{booktabs}
\usepackage{makecell}
\usepackage[table,dvipsnames]{xcolor}
\usepackage{float}

\title{Don’t Just Look, Intervene: Perturbation Based Region Labeling for VQA Images}

\author{
Marko Jojic \\
Arizona State University \\
\texttt{mjojic@asu.edu}
\And
Zhaonan Li \\
Arizona State University \\
\texttt{zhaonan2@asu.edu}
\AND
Ben Zhou \\
Arizona State University \\
\texttt{xzhou202@asu.edu}
}

\newcommand{\scorebase}[1]{%
  \makebox[3.0em][r]{#1}\hspace{0.10em}%
  \makebox[2.55em][l]{\textcolor{gray}{\scriptsize }}}

\newcommand{\scoreup}[2]{%
  \makebox[3.0em][r]{#1}\hspace{0.10em}%
  \makebox[2.55em][r]{\textcolor{ForestGreen}{\scriptsize (+#2)}}}

\newcommand{\scoredown}[2]{%
  \makebox[3.0em][r]{#1}\hspace{0.10em}%
  \makebox[2.55em][r]{\textcolor{BrickRed}{\scriptsize (-#2)}}}

\newcommand{\scoreupb}[2]{%
  \makebox[3.0em][r]{\textbf{#1}}\hspace{0.10em}%
  \makebox[2.55em][r]{\textcolor{ForestGreen}{\scriptsize (+#2)}}}

\newcommand{\scoreupu}[2]{%
  \makebox[3.0em][r]{\underline{#1}}\hspace{0.10em}%
  \makebox[2.55em][r]{\textcolor{ForestGreen}{\scriptsize (+#2)}}}

\newcommand{\scorebasedu}[1]{%
  \makebox[3.0em][r]{\underline{#1}}\hspace{0.10em}%
  \makebox[2.55em][l]{\textcolor{gray}{\scriptsize }}}

\definecolor{darkblue}{rgb}{0, 0, 0.5}
\hypersetup{colorlinks=true, citecolor=darkblue, linkcolor=darkblue, urlcolor=darkblue}

\newcommand{\sd}[1]{{\scriptsize$\pm$#1}}

\newcommand{\scorebasesd}[2]{\scorebase{#1}\,\sd{#2}}
\newcommand{\scorebasedusd}[2]{\scorebasedu{#1}\,\sd{#2}}

\newcommand{\scoreupsd}[3]{\scoreup{#1}{#2}\,\sd{#3}}
\newcommand{\scoreupusd}[3]{\scoreupu{#1}{#2}\,\sd{#3}}
\newcommand{\scoreupbsd}[3]{\scoreupb{#1}{#2}\,\sd{#3}}

\newcommand{\scoredownsd}[3]{\scoredown{#1}{#2}\,\sd{#3}}

\newcommand{\method}[0]{\textsc{CSGR}}
\newif\ifshowcomments
\showcommentstrue
\showcommentsfalse

\ifshowcomments
  \newcommand{\ben}[1]{\textcolor{purple}{[Ben: #1]}}
  \newcommand{\zhaonan}[1]{\textcolor{blue}{[Zhaonan: #1]}}
  \newcommand{\marko}[1]{\textcolor{red}{[Marko: #1]}}
\else
  \newcommand{\ben}[1]{}
  \newcommand{\zhaonan}[1]{}
  \newcommand{\marko}[1]{}
\fi

\begin{document}
\maketitle
\begin{abstract}
Vision Language Models (VLMs) should rely on visual evidence that directly determines the correct answer, but supervision for grounding visual reasoning is often expensive to obtain manually or tied to dataset-specific annotation primitives. We instead introduce \emph{model-causal visual evidence} as an annotation target, defined as the set of image regions whose counterfactual intervention changes a model's answer distribution for a given image-question pair. Based on this principle, we introduce Counterfactual Search for Grounding Regions (\method{}) \footnote{We will release all code and data under an open-source license upon publication.}. \method{} is a scalable pipeline that proposes candidate regions, perturbs them, measures their effect on answer sensitivity, and aggregates this evidence across multiple judges to approximate answer-critical regions in VQA data. To assess whether \method{} annotations contain a useful supervision signal, we plug them into three existing grounding-aware training routines: attention steering, Visual CoT finetuning, and latent visual reasoning. These experiments test whether the proposed annotation scheme can provide a useful supervision signal across multiple ways of consuming region labels, rather than introducing a new way of using them. Across competing automatic region-labeling mechanisms, \method{} annotations provide the most consistent gains over Cross Entropy-only finetuning in both in-domain and out-of-domain evaluations, indicating that the proposed labeling scheme captures useful region-level information. 
\end{abstract}

\section{Introduction}

\zhaonan{In the abstract and intro, we need to clarify our contribution is automatic labeling not a new training method. We also need to soften human vs model labeling distinction (better alignment with human judgement -> better labels). }

Robust visual reasoning requires more than predicting the correct answer: a model's prediction should be grounded in the visual evidence that actually determines it. Otherwise, a vision-language model (VLM) may succeed for the wrong reasons, relying on shortcuts, superficial correlations, or incomplete cues that break under distribution shift. A natural way to encourage this behavior is through \emph{visual grounding supervision} (VGS), which trains models not only to answer correctly but also to attend to the regions that support the answer. Prior work shows that supervising such visual evidence can improve both predictive performance and the faithfulness of model behavior \citep{hint, selfcritical, visfis}, suggesting that grounding supervision is a promising route to more robust visual reasoning. At the same time, its success depends critically on the quality of the supervision signal, raising a central question: what kind of visual evidence should grounding supervision target in the first place? \citep{reich-etal-2023-measuring} \zhaonan{if we adjust the narrative, need to remove or soften the question here.}


A common target for VGS is the visual evidence marked by a human as supporting the answer. Human-relevant evidence describes what people consider salient, whereas model-causal evidence describes what models are sensitive to. 
Moreover, collecting fine grained human annotations at scale is difficult, especially for reasoning-heavy datasets where the critical evidence is subtle and compositional.
Many automated annotation pipelines trade granularity for coverage, with large scale supervision coming from points rather than pixel-level evidence \citep{pixmo}. Existing methods often depend on dataset specific structures such as scene graphs, bounding boxes, or OCR regions, limiting their portability to new datasets \citep{shao2024visual}. As a result, current sources of supervision are limited by cost of human annotation and by the granularity, flexibility, and transferability of the signal from labeling pipelines.



More importantly, many existing supervision sources are only weakly aligned with the actual reasoning problem. They often annotate \emph{what is present in the image} rather than \emph{what should be attended to for answering a particular question}. For example, labeling the location of a person, object, or text region provides generic scene information, but not necessarily the evidence that is answer-critical for the current query. In contrast, visual reasoning requires \emph{question-dependent} supervision: the relevant region should change depending on what is being asked, and the amount of supervision should adapt to the complexity of the question. This distinction between generic object localization and question-conditioned, answer-relevant evidence is central to our work.

In this work, we propose an automatic annotation scheme for a \emph{model-causal} notion of visual evidence. We define a region's importance through \emph{counterfactual answer sensitivity}: given an image, question, and a model, we intervene on candidate regions and measure how the model's answer distribution changes. Regions whose perturbation causes a sufficiently large change in the predicted answer are treated as important, as the outcome of the model strongly depends on the visual information they contain. We refer to these regions as \emph{model-causal} visual evidence.
These regions are not intended to replace human answer-critical annotations. They follow a different labeling criterion: whether intervening on a region changes a model's prediction for the current image-question pair. They provide a measurable and scalable target based on a model's own sensitivity to these regions under intervention.

To automatically estimate this signal, we introduce Counterfactual Search for Grounding Regions (\method{}), a pipeline that proposes candidate regions using a segmentation model, perturbs them with an inpainting intervention model, and selects the regions with the strongest effect on answer prediction. Because importance estimates from a single judge model can be noisy or biased, \method{} aggregates evidence across multiple judge models through consensus to reduce dependence on any single judge. Unlike supervision based on points or coarse boxes, our method produces fine-grained pixel-level annotations. Moreover, because it is defined through question-conditioned counterfactual sensitivity rather than dataset-specific schemas, it transfers across datasets and reasoning settings.

Our primary contribution is the annotation scheme: a scalable way to collect model-causal visual grounding labels defined by counterfactual answer sensitivity. We demonstrate the use of \method{} labels on existing visual grounding training routines as probes of annotation utility. \method{} labels act as a supervision source for three methods of consuming region annotations: (\textit{i}) attention steering, (\textit{ii}) finetuning the \textsc{VisCoT} model, and (\textit{iii}) latent visual reasoning (\textsc{LVR}) fine-tuning. These experiments empirically show that \method{} labels provide useful supervision across these different mechanisms. Since our contribution is an automated labeling scheme for model-causal evidence, our primary comparison is between automatic annotation sources, with training settings as a controlled variable. Human annotations are useful as an auxiliary diagnostic of annotation behavior, but they are not the target of \method{}.

\zhaonan{This paragraph clarifies some misconceptions. Another point can be the improvement brought by VG is also method/model dependent, so they can't assume we consistantly have large improvements for all methods/models}
Our main comparison asks whether intervention-derived annotations provide a better automatic region-labeling source than existing automatic proposals. We therefore compare \method{} against alternative automatic label sources, including labels derived from pretrained grounding models such as \textsc{MOLMO2} and \textsc{VisCoT}. These results are measured with respect to Cross Entropy-only training without region labels. We additionally validate label quality through human-alignment and qualitative analyses, but do not treat human-label fine-tuning as the primary baseline.
Across all three settings, \textsc{CSGR} is the only automatic label source that consistently improves over Cross Entropy-only supervision, with the largest gains appearing under attention steering and \textsc{VisCoT}-style training. We interpret these results as evidence that the proposed annotation scheme captures useful model-causal region information, not as an optimal training routine using \method{} labels.

\section{Related Work}

\paragraph{VLM Robustness and Visual Grounding.} Literature on the robustness of VLMs has shown that strong VQA accuracy can coexist with weak visual grounding. For example, VLM accuracy drops sharply under rebalanced test sets or shifted answer priors, showing that models often exploit spurious correlations rather than robust visual evidence \citep{Goyal_2017_CVPR, dont_assume}. These results show that even when strong visual reasoning is expected in order to achieve high accuracy on a task, VLMs are likely to find other cues to rely on - yielding a brittle reasoning process. Recent benchmarks that assess robustness show that VQA systems often fail under controlled domain shifts, or when provided with natural adversarial examples that humans would not struggle with \citep{Li2022SuperCLEVRAV, NaturalBench}.

\paragraph{Grounding-aware Training with External Supervision.}
Vision language models tend to fall back on language priors under uncertainty, but explicitly tuning them to align model sensitivity with human performance can improve their visual grounding \citep{hint}. Prior work explores how region-level grounding signals can be incorporated into training for visual question answering and visual reasoning. Recent approaches have expanded this idea to richer training paradigms. Attention-based supervision methods regularize attention between text tokens and relevant image regions \citep{Qiao2017ExploringHA, guidingvisualattention}. Visual explanation through self critiquing can be used as a form of supervision as well, making correct answers more aligned with influential regions \citep{selfcritical}. Visual CoT uses image region annotations to isolate key visual evidence through cropping before producing a final answer \citep{shao2024visual}. Latent visual reasoning instead supervises reasoning through intermediate visual tokens that are produced during the reasoning process \citep{li2026latent}. Weakly supervised grounding follows a similar pattern, clustering features into groups for classification \citep{Khan2022WeaklySG}. 

\paragraph{Visual Grounding Label Sources.}
Existing grounding supervision typically comes from one of several external sources. Some work relies on human annotations, such as human attention maps or manually marked regions associated with a question-answer pair. These annotations can provide useful and interpretable supervision, but they are costly to collect at scale and primarily capture human-relevant evidence rather than the evidence a model’s prediction is actually sensitive to \citep{human_attention}. Other approaches repurpose annotation structures already present in a dataset, using them as region-level supervision during training by linking segmentations to VQA question-answer pairs \citep{linkingsegmentation}. A third class of methods uses fine tuned models to output heuristic or proposal-based mechanisms, such as prompted bounding boxes in \textsc{VisCoT} \citep{shao2024visual} or point-based region proposals in \textsc{MOLMO2}, to approximate grounding labels when explicit annotations are unavailable \citep{pixmo}. In the case of \textsc{VisCoT}, many bounding boxes in its training data are inherited from annotation structures such as OCR, which can localize the answer-bearing region without capturing the broader visual context whose joint presence makes the answer correct. \textsc{pixmo}, on the other hand, is trained to point to the target concept specified in the prompt by selecting the corresponding visual tokens, rather than to recover a question-specific region validated by its effect on the final answer. In contrast, \method{} defines the label target through intervention: a region is selected when perturbing it changes a judge model's answer distribution for the current image-question pair. Thus, the distinction is not merely that our labels are automatic, but that they estimate a different causal target.


\paragraph{Faithfulness and Counterfactual Views of Grounding.}
A smaller but important line of research asks not only whether a predicted answer is correct, but whether the model is \emph{faithfully grounded} in the relevant visual evidence. Prior work has shown that improvements in grounding or attention quality do not necessarily imply that a model is relying on the right evidence for the right reasons \citep{shrestha-etal-2020-negative}. Along this line, plausible visual grounding can be measured by comparing model behavior when provided relevant-only against irrelevant-only subsets of objects \citep{reich-etal-2023-measuring}. Counterfactual editing of images and questions has been used to combat bias during training, including masking critical objects and assigning appropriately changed answers \citep{counterfactual_samples}. Some recent methods use perturbations to probe model reliance on particular image regions for analysis and inference time gains \citep{wan2024contrastiveregionguidanceimproving, Wu2025AntidoteAU}. Swapping visual features between images in a dataset is a similar approach that can be used to diagnose an over-reliance on irrelevant visual context \citep{swapmix}. Much like the counterfactual perturbation of images, this approach can be used as a form of regularization through data augmentation during training. Our work aligns most closely with this research trajectory. However, rather than using interventions only for analysis, inference, or auxiliary training objectives, we define \emph{model-causal visual evidence} itself as the supervision target and introduce a pipeline that can recover labels for that target.




\section{Methodology}

\begin{figure*}[t]
    \centering
    \includegraphics[
        width=0.95\textwidth,
        clip,
        trim={1.8cm 5cm 1.9cm 4cm}
    ]{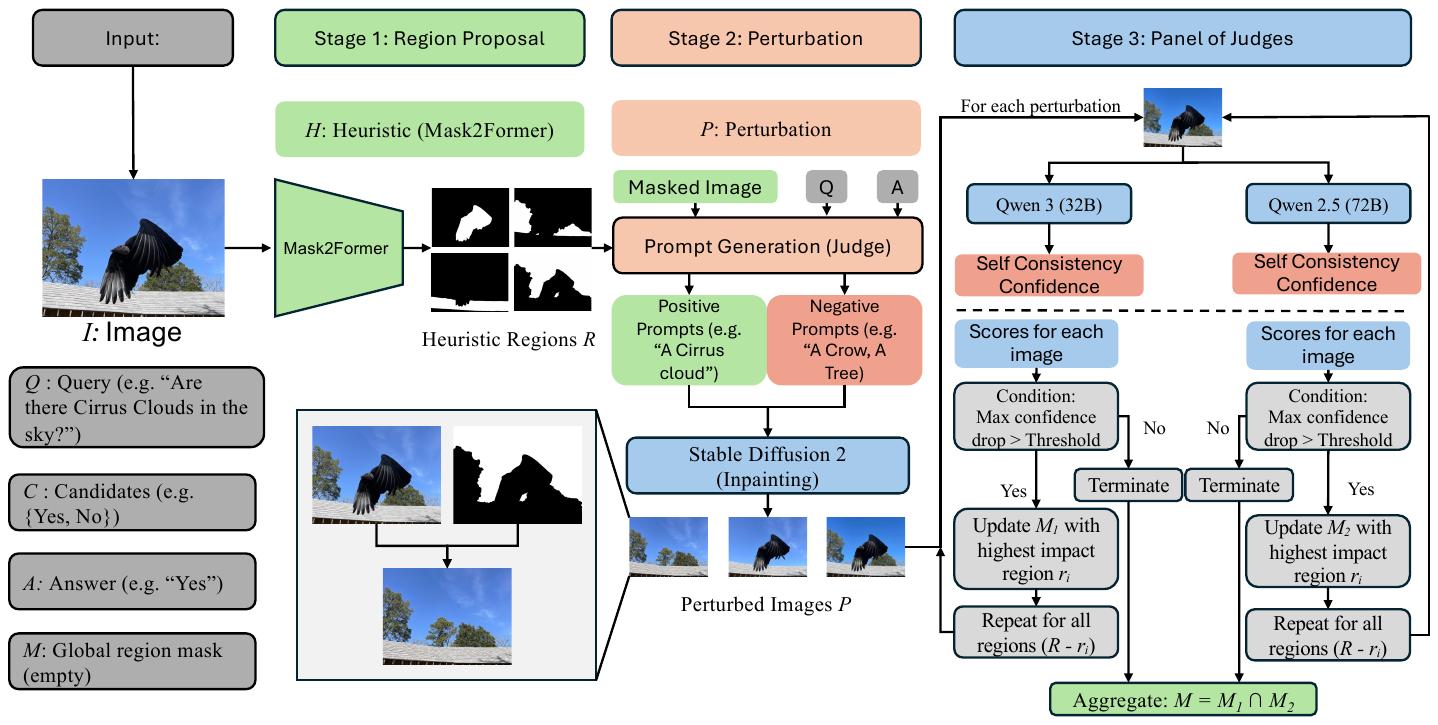}
    \caption{\method{} pipeline for finding model-causal image regions}
    \label{fig:methodology}
\end{figure*}




\subsection{Problem Setup}
Given a VQA instance and a VLM, $(\mathcal{I}, Q, C, A, \theta)$, where $\mathcal{I}$ is an image, $Q$ is a question about the image, $C$ is a set of candidate answers, $A \in C$ is the gold answer, and $\theta$ denotes the model, our goal is to produce a pixel-level mask $M$ identifying the decision-critical region $G \subseteq \mathcal{I}$. Informally, $G$ contains the visual evidence whose removal or perturbation reduces the model's confidence over the gold answer $A$. Concretely, if $\mathcal{I}_p$ denotes the perturbed image obtained by intervening on the masked region, then perturbing $G$ should substantially reduce the model's support for the correct answer:
\[
p(A \mid \theta, Q, C, \mathcal{I}_p) \ll p(A \mid \theta, Q, C, \mathcal{I}).
\]
Our objective is to recover a mask $M$ that approximates this decision-critical region.

\subsection{Global view of the solution}

Our pipeline builds on three main components: $H$ - a heuristic, or region proposer, $\Theta$ - a set of VLMs, used to judge regions, $P$ - a region perturbation procedure.
Together $H$ and $P$ provide perturbed versions of the original image. We can measure the importance of a region based on how a VLM's confidence in the gold answer changes when given the perturbed image $P(H(I)_i)$, where $H(I)_i$ indicates the $i$-th region proposed by the heuristic. This is measured as $SC(A | \theta, Q, C, I) - SC(A | \theta, Q, C, P(H(I)_i))$, where $SC$ represents the answer likelihood based on Self Consistency as a confidence metric. 
The process of a judge selecting the most important region is repeated until there are no remaining candidate regions that cause a drop in the gold answer's likelihood when perturbed. CSGR then aggregates the region selection of each judge through a consensus; in our main two judge setting, we take the intersection of judge-specific masks. The single-judge search procedure is given in Algorithm~\ref{alg:csgr_single}.

\subsection{Heuristic}

The region proposer gives us the set of candidate regions - determining the search space which our pipeline explores. In our experiments, we use Mask2Former, a segmentation model introduced by \cite{mask2former}. This model outputs a set of pixel-level objects for each image (average 7 objects on NaturalBench), giving us a fine grained signal to build on.



\subsection{Perturbation}

As a perturbation scheme, we use Stable Diffusion 2
for inpainting candidate regions of the image \footnote{\url{https://huggingface.co/docs/diffusers/api/pipelines/stable_diffusion/stable_diffusion_2}}. This choice was made due to implicit side effects of other approaches (filling in a region with color or a blur). Applying a blur can still retain the underlying semantic of that region, and coloring a region injects a new semantic that may confuse the judge model. For each image region, we obtain a unique positive/negative prompt to send to the inpainting model. This is done by sending a judge the image (with the region masked) alongside the question and gold answer - prompting it to return a list of items that should/shouldn't appear in the image in order to make the gold answer false.

\subsection{Panel of judges}

By using a VLM to judge the importance of a patch (likelihood of the gold answer given the model's parameters, perturbed image, and question), the result of our search is dependent on the inductive biases of the model. We mitigate this issue by performing the \method{} search using multiple judge VLMs to produce a label for a given image and selecting only the intersection of the judge-proposed regions. The aggregated region is then validated through perturbation, requiring that its intervention is still sufficient to change both judges' answers. In our experiments, we use Qwen3-VL 32B and Qwen 2.5 VL 72B as two judges.

\section{Experiments}

Our experiments are focused on answering three questions: 
\begin{enumerate}
    \item Do \method{} labels provide useful in-domain region supervision beyond answer-only Cross Entropy (VQA loss)?
    \item Do approximations of model-causal evidence transfer better than alternative automatic region labels under distribution shift?
    \item Is the utility of the annotations visible across multiple ways of using region labels, rather than being specific to a single training routine?
\end{enumerate}


Our evaluation focuses on automatic sources of region supervision rather than human annotations because the target of \method{} is not human relevance. A region can be human-relevant without being causally used by a model, and a model can be sensitive to regions that are not the regions a human would choose. We therefore compare \method{} against automatic region-labeling mechanisms that can be used as supervision sources in the same training pipelines. We choose our baselines from models that have been trained on massive VG annotations, including \textsc{VisCoT} and \textsc{MOLMO2}. To isolate the gain from VG, we measure delta accuracy compared against training without VG (VQA loss only).

We evaluate the proposed labels by using them as supervision signals in existing visual-grounding training routines. Each routine consumes region supervision differently: attention steering uses labels as an attention regularizer, \textsc{LVR} uses them to supervise latent visual token reconstruction, and \textsc{VisCoT} supervises prediction of the labeled regions for cropping.



We employ \method{} labels across three training paradigms: Visual CoT (\textsc{VisCoT}), Latent Visual Reasoning (LVR), and Attention Steering. We compare models trained using our labels with four baselines: training with Cross-Entropy only, bounding boxes produced by \textsc{VisCoT}, bounding boxes produced by Qwen3-VL 235B, and bounding boxes produced by \textsc{MOLMO2}'s point detection pipeline. The resulting models are scored on four datasets: NaturalBench held out images, MMVP, POPE, and BLINK. For each score, we report the average accuracy over 3 separate training seeds. 

\subsection{Label Extraction}

We employ \method{} as described in section 3 on 5000 samples from the NaturalBench dataset of which we found 2019 labels such that perturbing those regions causes both judges to decrease their confidence past a threshold. To the baseline models, we provide the image, question, and gold answer - prompting \textsc{VisCoT}, Qwen3-VL 235B and \textsc{MOLMO2} 8b models to produce bounding boxes they identify as relevant to answering the question. \textsc{VisCoT} and Qwen3-VL 235B are prompted to output a bounding box as a tuple: $(x_{low}, y_{low}, x_{high}, y_{high})$ with values scaled between 0-1. \textsc{MOLMO2}, however, is trained to identify points in an image, so we query it to produce four points representing the corners of a bounding box around the critical region. These baselines produce plausible grounding regions, but unlike \method{}, they are not selected by measuring the counterfactual effect of a region's perturbation on a model's answer distribution. We treat all four sources as alternative mechanisms for producing region annotations over the same image-question-answer instances. The downstream training experiments then compare the utility of these annotations when plugged into identical training routines.

\zhaonan{add more details in appendix? e.g., how do we prompt the judge, \textsc{MOLMO2}, Qwen235B, etc.}

\subsection{Training method: Attention Steering}

\zhaonan{to save space, we need to move the loss function details to appendix. just describe on a high-level what it does}

For this experiment, we train the Qwen3-VL 8B model on our labeled data using the attention steering strategy, prompted to output a single answer token without chain of thought. The primary component of the loss function is Cross Entropy over the answer choices, along with an alignment loss. To measure the attention alignment, we extract the attention weights between the answer token and visual tokens in the final layer of the model. We then measure the distance between this attention map and a token level representation of our label. This score is added to the Cross Entropy loss, multiplied by a scalar hyper parameter representing the sensitivity to the visual supervision.

Let $x = (I, q)$ denote an image-question pair, let $y^\star$ be the correct answer token, let $A \in \mathcal{R}^{T_v}$ be the attention in the final transformer layer between the answer tokens and visual tokens, and $M \in \{0,1\}^{T_v}$ be the token-level grounding label over the $T_v$ visual tokens, where
\[
M_j =
\begin{cases}
1 & \text{if visual token } j \text{ in the labeled region,} \\
0 & \text{otherwise.}
\end{cases}
\]

The answer prediction loss is the standard cross-entropy objective:
\[
\mathcal{L}_{\mathrm{CE}} = -\log p_\theta(y^\star \mid x).
\]

To encourage visual grounding, we compare $A$ to the token-level grounding label $M$ using an alignment loss
where $SSE(\cdot,\cdot)$ is the sum of squared error between the attention map and the binary grounding mask.

The full training objective is then:
\[
\mathcal{L}
=
\mathcal{L}_{\mathrm{CE}}
+
\lambda SSE(A, M)
\]
\zhaonan{ just $\frac{1}{|T_v|}||A - M||_2^2$}
where $\lambda > 0$ controls the strength of the grounding supervision. In our experiments, we initialize $\lambda$ such that $\lambda SSE(A, M)$ is roughly a quarter of the magnitude of $\mathcal{L}_{\mathrm{CE}}$. $\lambda$ is tuned for each label source based on early epoch performance.

\begin{table*}[t]
\centering
\small
\setlength{\tabcolsep}{4pt}
\renewcommand{\arraystretch}{1.15}
\begin{tabular}{lcccc}
\toprule
& \textbf{In-Domain} & \multicolumn{3}{c}{\textbf{Out-of-Domain}} \\
\cmidrule(lr){2-2} \cmidrule(lr){3-5}
\textbf{Supervision} & \textbf{NB-Val} & \textbf{MMVP} & \textbf{POPE} & \textbf{BLINK} \\
\midrule

\multicolumn{5}{l}{\textbf{Attention Steering}} \\
\rowcolor{gray!8}
CE only
& \scorebasesd{82.33}{0.24}
& \scorebasedusd{80.33}{0.13}
& \scorebasedusd{88.33}{0.09}
& \scorebasesd{68.33}{0.45} \\
CE + VG (\textsc{VisCoT})
& \scoredownsd{80.84}{1.49}{0.32}
& \scoredownsd{79.67}{0.66}{0.18}
& \scoredownsd{87.78}{0.55}{0.39}
& \scoredownsd{65.83}{2.50}{0.51} \\
CE + VG (Qwen3-VL-235B)
& \scoredownsd{81.02}{1.31}{0.65}
& \scoredownsd{79.33}{1.00}{0.38}
& \scoredownsd{87.83}{0.50}{0.47}
& \scoredownsd{66.67}{1.66}{0.29} \\
CE + VG (\textsc{MOLMO2})
& \scoreupusd{83.02}{0.69}{0.28}
& \scoredownsd{80.00}{0.33}{0.09}
& \scoredownsd{87.97}{0.36}{0.21}
& \scoreupusd{69.17}{0.84}{0.43} \\
CE + VG (\method{}, ours)
& \scoreupbsd{85.80}{3.47}{0.76}
& \scoreupbsd{80.67}{0.34}{0.41}
& \scoreupbsd{88.71}{0.38}{0.25}
& \scoreupbsd{70.83}{2.50}{0.32} \\

\addlinespace[2pt]
\multicolumn{5}{l}{\textbf{LVR}} \\
\rowcolor{gray!8}
CE only
& \scorebasedusd{79.60}{0.03}
& \scorebasedusd{71.44}{0.19}
& \scorebasesd{87.41}{0.05}
& \scorebasedusd{69.45}{0.09} \\
CE + VG (\textsc{VisCoT})
& \scoredownsd{75.69}{3.91}{0.05}
& \scoredownsd{70.67}{0.77}{0.12}
& \scoredownsd{87.31}{0.10}{0.08}
& \scoredownsd{68.05}{1.40}{0.05} \\
CE + VG (Qwen3-VL-235B)
& \scoredownsd{75.79}{3.81}{0.04}
& \scoredownsd{70.78}{0.66}{0.09}
& \scoredownsd{87.40}{0.01}{0.07}
& \scoredownsd{68.61}{0.84}{0.19} \\
CE + VG (\textsc{MOLMO2})
& \scoredownsd{78.10}{1.50}{0.13}
& \scoredownsd{71.33}{0.11}{0.34}
& \scoreupusd{87.44}{0.03}{0.12}
& \scoredownsd{68.89}{0.56}{0.11} \\
CE + VG (\method{}, ours)
& \scoreupbsd{79.77}{0.17}{0.02}
& \scoreupbsd{72.22}{0.78}{0.04}
& \scoreupbsd{87.53}{0.12}{0.07}
& \scoreupbsd{69.72}{0.27}{0.25} \\

\addlinespace[2pt]
\multicolumn{5}{l}{\textbf{\textsc{VisCoT}}} \\
\rowcolor{gray!8}
CE only
& \scorebasesd{74.64}{0.26}
& \scorebasesd{62.00}{0.53}
& \scorebasedusd{83.93}{0.3}
& \scorebasedusd{46.67}{0.18} \\
CE + VG (\textsc{VisCoT})
& \scoredownsd{72.98}{1.66}{0.42}
& \scoreupsd{62.11}{0.11}{0.29}
& \scoredownsd{82.84}{1.09}{0.15}
& \scoredownsd{45.00}{1.67}{0.51} \\
CE + VG (Qwen3-VL-235B)
& \scoredownsd{73.02}{1.62}{0.32}
& \scoreupsd{62.22}{0.22}{0.26}
& \scoredownsd{82.95}{0.98}{0.19}
& \scoredownsd{45.83}{0.84}{0.24} \\
CE + VG (\textsc{MOLMO2})
& \scoreupusd{75.67}{1.03}{0.43}
& \scoreupusd{63.00}{1.00}{0.14}
& \scoredownsd{83.17}{0.76}{0.27}
& \scoreupusd{46.67}{0.00}{0.35} \\
CE + VG (\method{}, ours)
& \scoreupbsd{75.78}{1.14}{0.52}
& \scoreupbsd{63.55}{1.55}{0.29}
& \scoreupbsd{84.15}{0.22}{0.38}
& \scoreupbsd{47.50}{0.83}{0.46} \\

\bottomrule
\end{tabular}
\caption{Comparison across three ways of applying important-region supervision: \textbf{Attention Steering}, \textbf{LVR}, and \textbf{\textsc{VisCoT} Training}. Within each block, \textit{CE only} is the baseline using standard VQA cross-entropy loss without visual grounding (VG) supervision. All other rows add VG supervision on top of the same CE objective, differing only in the source of important-region labels. Colored deltas are computed relative to the CE-only baseline within each block. Bold and underline mark the best and second-best results within each block, respectively. Each cell reports mean accuracy with standard deviation over 3 training seeds.}
\label{tab:merged_results}
\end{table*}

In the attention steering section of table \ref{tab:merged_results}, we see that only the labels from \method{} and \textsc{MOLMO2} can outperform the Cross Entropy only training baseline. Of these two approaches, \method{} shows a much stronger gain on the in domain evaluation as well as the BLINK dataset. This result is accompanied by some smaller gains over CE only training on the MMVP and POPE evaluations. These results highlight the utility of \method{} labels for in domain tasks, as well as a degree of transferability to OOD settings.

\subsection{Training Method: Latent Visual Reasoning (LVR)}

LVR's training pipeline contains an SFT stage and a GRPO stage. We focus only on the SFT stage due to the computational costs associated with their GRPO, mentioned in their paper \citep{li2026latent}. The SFT stage contains a standard VQA loss alongside a reconstruction loss between the intermediate visual tokens and the visual tokens representing the labeled region(s) of the input image. For these experiments we use the LVR model with Qwen 2.5 VL 7B as a backbone to remain close in size to the baselines. At inference time the number of intermediate visual tokens is fixed to 8 (from default options 1, 2, 4, 8, 16), observed as a point of diminishing returns on the NaturalBench dataset.

\[
\mathcal{L}_{\text{LVR}}
=
\mathcal{L}_{\text{CE}} + \lambda \mathcal{L}_{\text{ground}},
\]
where
\[
\mathcal{L}_{\text{ground}}
=
\frac{1}{T_v}\sum_{t=1}^{T_v}
\left\| z_t - v_t^{\text{ROI}} \right\|_2^2.
\]
Here, \(z_{1:T_v}\) are the latent visual reasoning states, and \(v_{1:T_v}^{\text{ROI}}\) are the visual tokens of the labeled region(s).

In table \ref{tab:merged_results}, we observe that in the LVR framework, \method{} labels show a small improvement over the CE only training, with the largest gain on MMVP. However, \textsc{VisCoT}, Qwen 235B, and \textsc{MOLMO2} show reduced accuracy compared to the baseline across the board (outside of \textsc{MOLMO2}'s small gain on POPE). While the margin is slimmer than in attention steering, this is still encouraging evidence that the \method{} labels provide a stronger signal compared with other sources.

\subsection{Training method: Visual CoT (\textsc{VisCoT})}

The \textsc{VisCoT} inference pipeline first prompts the VLM to output a bounding box corresponding to the relevant region of the image. It then crops that region of the image and feeds it back to the \textsc{VisCoT} model alongside the original image when prompting for the final answer. During training, the crop is teacher-forced using the ground truth bounding box, rather than from the \textsc{VisCoT} output. The model is supervised with Cross Entropy over the answer sequence (bounding box + optional annotated CoT + answer tokens). When bounding boxes are omitted, there is no cropping, and only the text tokens are supervised. In order to use their training code with \method{} labels, we convert our pixel-level masks to bounding boxes.

\[
\begin{aligned}
\mathcal{L}_{\text{VisCoT}}(s)
&= -\frac{1}{|s|}\sum_{t=1}^{|s|}
\log p_{\theta}(s_t \mid s_{<t}, I, I_{\mathrm{crop}}),\\
\mathcal{L}_{\text{VisCoT}}(a)
&= -\frac{1}{|a|}\sum_{t=1}^{|a|}
\log p_{\theta}(a_t \mid a_{<t}, I).
\end{aligned}
\]
Here, \(I_{\mathrm{crop}}\) is extracted from the teacher-forced bounding box \(b^\star\). 
The supervised sequence is \(s=[b^\star;r;a]\) when reasoning text is available, 
\(s=[b^\star;a]\) when only a bounding box is provided, and \(s=[a]\) when no region annotation is provided.

\[
s =
\begin{cases}
[b^{\star};\, r;\, a] & \text{with reasoning},\\
[b^{\star};\, a] & \text{box only},\\
[a] & \text{no region label}.
\end{cases}
\]

Visual alignment is important in the \textsc{VisCoT} framework - directly influencing the final inputs when generating an answer at inference time. Thus, table \ref{tab:merged_results} tells a slightly different story on the \textsc{VisCoT} training evaluation, although following the same general trend as the previous experiments. All four of the labeling approaches provided some gain in the OOD evaluation on MMVP, but \textsc{MOLMO2} and \method{} showed the strongest results across all evaluations. On MMVP, POPE, and BLINK, \method{} consistently outperforms \textsc{MOLMO2} supervision and the CE only baseline.

\section{Ablations and Analysis}

\subsection{Labeling SeedBench}

To test whether \method{}'s in-domain gains generalize to other datasets, we collected 1055 region labels for a subset of 3000 images in the SeedBench dataset. With our previous experiments showing \textsc{MOLMO2} as our strongest baseline, we compare only against this label source. As attention steering showed the strongest results in the previous experiments, we selected this as the training method. We evaluate the accuracy on the held out images using the models learned from both label sources.

\begin{table}[h]
\centering
\small
\begin{tabular}{lc}
\toprule
\textbf{Model / Training Setup} & \textbf{SeedBench Avg Acc} \\
\midrule
Qwen base & 79.02 \\
Qwen + VQA & 81.60 \\
Qwen + VQA + AS (\textsc{MOLMO2}) & \underline{82.84} \\
Qwen + VQA + AS (\method{}) & \textbf{84.28} \\
\bottomrule
\end{tabular}
\caption{Comparison between base Qwen, finetuning with VQA loss, and finetuning with attention steering, supervised by labels mined from SeedBench data}
\label{tab:SeedBench_attn}
\end{table}

Table \ref{tab:SeedBench_attn} evaluates whether the in distribution gains from model-causal labeling persist on a different dataset. All trained variants improve over the base model, with VQA finetuning alone providing a substantial gain. Adding grounding supervision further improves performance, indicating that region-level signals remain useful in this setting.
\method{} labels outperform VQA finetuning and \textsc{MOLMO2} supervision. Consistent improvement across datasets shows that model-causal labels provide strong in-domain supervision signals that translate into improved task performance.

\subsection{\method{} Applied Across Model Families}
Our previous experiments use the same model family (Qwen) as judges and downstream recipients of the training signal produced by \method{}. This leaves the question of whether \method{} labels act as a useful supervision source for models outside of the judge family. To answer this question, we train Gemma 3 4B on NaturalBench, using the \method{} labels mined using Qwen 3 32B and Qwen 2.5 72B as judges. We use attention steering to enforce the visual grounding objective, comparing against the base Gemma model and one finetuned with a VQA loss.

\begin{table}[h]
\centering
\small
\setlength{\tabcolsep}{8pt}
\begin{tabular}{lcc}
\toprule
Model & Accuracy (\%) & Std. Dev. \\
\midrule
Gemma Base & 67.75 & 0.47 \\
Gemma + VQA & 74.26 & 0.94 \\
Gemma + AS (\method{}) & 77.78 & 0.47 \\
\bottomrule
\end{tabular}
\caption{NaturalBench accuracy of Gemma after fine-tuning with different training strategies. Results are averaged over three independent runs.}
\label{tab:gemma_results}
\end{table}

\zhaonan{additional results to be considered:
\begin{enumerate}
    \item attention map before and after steering
    \item quantified human judgment alignment 
    \item a discussion on time/compute used for obtaining our labels (appendix)
\end{enumerate}
}

\subsection{Additional Analysis}

Our main experiment uses two judges, Qwen3-VL 32B and Qwen 2.5 VL 72B, and accepts a mask only under strict consensus. This makes label acceptance conservative: CSGR labels roughly 40\% of the dataset. To test whether coverage improves with a less restrictive aggregation rule, we rerun CSGR on 100 image-question pairs with larger judge pools. As shown in Table~\ref{tab:judge_coverage}, moving from two-judge consensus to majority vote with three or five judges substantially increases coverage, from 45 to 60 and 71/100 respectively. This suggests that the lower coverage in our main experiments is partly due to the strict consensus rule, not a fundamental limitation of the annotation scheme.

\begin{table}[h]
\centering
\small
\setlength{\tabcolsep}{6pt}
\begin{tabular}{lcc}
\toprule
\textbf{Judge pool} & \textbf{Labeled} \\
\midrule
Qwen3-32B, Qwen2.5-72B consensus & 45/100 \\
+ Llama 3.2 Vision 11B majority & 60/100 \\
+ InternVL3-14B, InternVL3-38B majority & 71/100 \\
\bottomrule
\end{tabular}
\caption{Label coverage increases as CSGR uses larger judge pools.}
\label{tab:judge_coverage}
\end{table}

While \method{} is designed to recover model-causal evidence rather than human relevance, we use alignment to human labels as a complementary diagnostic to downstream accuracy. We collect 50 image-question pairs for which all CSGR configurations produced labels, manually annotate region masks, and compute IoU between the human masks and each label source. As shown in Table~\ref{tab:human_iou}, \method{} masks have a higher overlap with human-marked regions than prompted bounding-box baselines. Adding more judges slightly increases mean overlap, suggesting that majority-vote aggregation with larger judge pools may improve alignment with human-marked regions.

\begin{table}[h]
\centering
\small
\setlength{\tabcolsep}{6pt}
\begin{tabular}{lcc}
\toprule
\textbf{Source} & \textbf{Mean IoU} & \textbf{S.D. IoU} \\
\midrule
\method{} 2 judges & 0.7400 & 0.2096 \\
\method{} 3 judges & 0.7563 & 0.2113 \\
\method{} 5 judges & \textbf{0.7694} & \textbf{0.1982} \\
\textsc{VisCoT} & 0.2017 & 0.2208 \\
Qwen3-VL 235B & 0.3846 & 0.2951 \\
\textsc{MOLMO2} & 0.6252 & 0.2647 \\
\bottomrule
\end{tabular}
\caption{IoU between human-marked answer regions and automatic region labels on 50 image-question pairs where all \method{} judge configurations produced labels. Human masks are used here as a diagnostic lens for characterizing annotation behavior, not as the optimization target of \method{}.}
\label{tab:human_iou}
\end{table}

\section{Conclusion}


In this paper we introduce model-causal evidence as a supervision target for visual grounding: image regions whose intervention changes a model's answer distribution for a specific question. \method{} provides a portable and scalable mechanism to recover this target without manual annotation or dataset-specific primitives. Across three training paradigms, \method{} labels provide consistent improvements over CE-only training and competing automatic label sources. These gains extend to out-of-domain datasets, highlighting the transferability of this signal. Our results suggest that model-causal evidence is a useful supervision target for grounding-aware VLM training. Rather than replacing human annotations, \method{} estimates which image regions a model's answer is sensitive to under intervention, enabling training toward decision-critical visual evidence.

\section{Limitations}
\method{} does not always find a model-causal mask for each image in the dataset - with two judges, we achieve roughly 40\% coverage. If there is enough of a disagreement between the judges, CSGR may not be able to find a consensus region to which each judge is sensitive. Additionally, if a perturbation of any single image region is insufficient to shift a judge model's answer distribution, CSGR will terminate after searching for the first region. However, we find in table \ref{tab:judge_coverage} that increasing the number of judges and using majority vote as a consensus can yield much better coverage.

Although \method{} is designed to recover model-causal evidence, its labels may sometimes expose artifacts of the region proposal and perturbation processes rather than the most semantically correct evidence for an answer as seen in figure \ref{fig:failure_cases}. For example, in questions about whether an object casts a shadow, \method{} may identify the object as the only critical region instead of the shadow itself, even though both are important in determining the answer.  Similarly, on counting tasks, \method{} may highlight the background of the image rather than the target objects, because intervening on this space can enable the addition of new objects to change the judges' answer distributions.

\newpage

\newpage

\bibliography{custom}

\clearpage

\appendix

\section{Appendix}

\subsection{Algorithms}

\begin{algorithm}[H]
\small
\caption{\method{}: Single-Judge Search}
\label{alg:csgr_single}
\begin{algorithmic}[1]
\Require Image $I$, question $Q$, answers $C$, gold answer $A$
\Require Region proposer $H$, judge $\theta$, perturbation $P$
\Require Threshold $\tau$, samples $N$
\Ensure Judge-specific mask $M_\theta$

\State $\mathcal{R} \gets H(I)$ \Comment{candidate regions}
\State $M_{\theta} \gets \emptyset$

\While{\textbf{true}}
    \State $r^\star \gets \texttt{None}$; $\Delta^\star \gets 0$

    \ForAll{$r \in \mathcal{R} \setminus M_\theta$}
        \State $\Delta_r \gets$
        \Statex \hspace{\algorithmicindent}
        $\textsc{JudgeScore}(\theta,P,I,Q,C,A,M_\theta,r,N)$
        \If{$\Delta_r > \Delta^\star$}
            \State $\Delta^\star \gets \Delta_r$; $r^\star \gets r$
        \EndIf
    \EndFor

    \If{$\Delta^\star < \tau$}
        \State \textbf{break}
    \EndIf

    \State $M_{\theta} \gets M_{\theta} \cup \{r^\star\}$
\EndWhile

\State \Return $M_\theta$
\end{algorithmic}
\end{algorithm}

\begin{algorithm}[H]
\small
\caption{\textsc{JudgeScore}}
\label{alg:judge_score_single}
\begin{algorithmic}[1]
\Require Judge $\theta$, perturbation $P$, image $I$, question $Q$
\Require Answers $C$, gold answer $A$, selected regions $S$
\Require Candidate region $r$, samples $N$
\Ensure Gold-answer support difference $\Delta$

\State $I_S \gets P(I, S)$
\State $I_{S \cup r} \gets P(I, S \cup \{r\})$

\State $s_{\text{base}} \gets \textsc{SC}(\theta, I_S, Q, C, A, N)$
\State $s_{\text{cand}} \gets \textsc{SC}(\theta, I_{S \cup r}, Q, C, A, N)$

\State \Return $s_{\text{base}} - s_{\text{cand}}$
\end{algorithmic}
\end{algorithm}

\begin{algorithm}[H]
\small
\caption{\method{}: Multi-Judge Greedy Search}
\label{alg:csgr_multi}
\begin{algorithmic}[1]
\Require Image $I$, question $Q$, answers $C$, gold answer $A$
\Require Region proposer $H$, judges $\Theta$, perturbation $P$
\Require Threshold $\tau$, samples $N$
\Ensure Aggregated mask $M$

\State $M \gets \texttt{None}$

\ForAll{$\theta \in \Theta$}
    \State $M_\theta \gets$
    \Statex \hspace{\algorithmicindent}
    $\textsc{SingleJudgeSearch}(I,Q,C,A,H,\theta,P,\tau,N)$
    \If{$M = \texttt{None}$}
        \State $M \gets M_\theta$
    \Else
        \State $M \gets M \cap M_\theta$
    \EndIf
\EndFor

\State $I_M \gets P(I,M)$
\ForAll{$\theta \in \Theta$}
    \State $s_{\text{base}} \gets \textsc{SC}(\theta,I,Q,C,A,N)$
    \State $s_{\text{agg}} \gets \textsc{SC}(\theta,I_M,Q,C,A,N)$
    \If{$s_{\text{base}} - s_{\text{agg}} < \tau$}
        \State \Return $\emptyset$
    \EndIf
\EndFor

\State \Return $M$
\end{algorithmic}
\end{algorithm}

\subsection{Computational Cost of CSGR}
The computational cost of CSGR is determined by the number of segmented regions in an image. At iteration $i$, the algorithm evaluates $S-i$ perturbed images, where $S$ is the number of candidate segments. With a maximum of $P \leq S$ selected regions, the worst-case number of perturbation evaluations (assuming no early stopping) is $\sum_{i=0}^{P}(S-i)$, yielding a worst-case computational complexity of $O(PS)$. On the NaturalBench dataset, Mask2Former produced on average seven candidate segments per image, resulting in an average of 47 LLM calls across both judge models using unique perturbed images. 

While CSGR incurs inference-time cost during annotation, it avoids training a dedicated grounding model. For comparison, VisCoT was trained on approximately 2.5 million instruction-tuning examples using 8 A100 80GB GPUs, while MOLMO2 reports training on 6.5 million examples requiring 2.27k H100 GPU-hours. Although inference and training costs are not directly comparable, CSGR trades inference costs at annotation time to avoid a substantially larger one-time cost of training a specialized labeling model.

\subsection{Statistical Significance of \method{} results}
While the results of \method{} in table~\ref{tab:merged_results} are strong with attention steering as a grounding strategy, \method{} showed much closer results to MOLMO2 labels when training with LVR or VisCoT strategies. To assess statistical significance of these improvements, we conduct one-sided paired t-tests comparing models trained with CSGR labels against those trained with MOLMO2 labels. For each benchmark, we tested the null hypothesis $H_0: \mu_{\text{CSGR}} \leq \mu_{\text{MOLMO2}}$ against the alternative hypothesis $H_1: \mu_{\text{CSGR}} > \mu_{\text{MOLMO2}}$, where $\mu$ denotes the mean accuracy. We report the resulting $p$-values in Table~\ref{tab:paired_t_tests}.

\begin{table}[h]
\centering
\small
\setlength{\tabcolsep}{8pt}
\begin{tabular}{lcccc}
\toprule
Method & NB & MMVP & POPE & BLINK \\
\midrule
LVR     & 0.0027 & 0.0428 & 0.0503 & 0.0001 \\
VisCoT  & 0.0031 & 0.0506 & 0.0395 & 0.00001 \\
\bottomrule
\end{tabular}
\caption{One-sided paired $t$-test $p$-values comparing models trained using CSGR labels against those trained using MOLMO2 labels. Lower $p$-values indicate stronger evidence that CSGR outperforms the MOLMO2 baseline.}
\label{tab:paired_t_tests}
\end{table}

\subsection{Comparing CSGR With and Without Aggregation}

\begin{figure*}[t]
    \centering
    \setlength{\tabcolsep}{2pt}
    \renewcommand{\arraystretch}{1.1}

    \begin{tabular}{ccc}
        \small \textbf{Qwen 3 mask} & \small \textbf{Qwen 2.5 mask} & \small \textbf{Consensus mask}
    \end{tabular}

    \vspace{2mm}

    \parbox{0.98\linewidth}{\centering \small \textbf{Question 1:} Is there a woman wearing red clothing in the picture?}
    
    \vspace{1mm}
    
    \begin{tabular}{ccc}
        \includegraphics[width=0.25\linewidth]{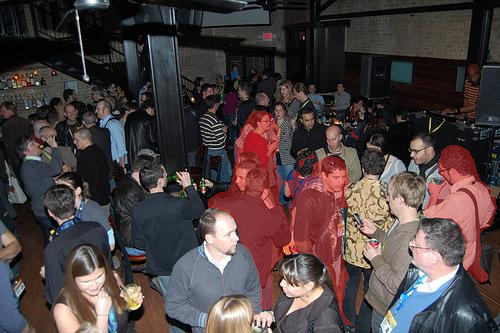} &
        \includegraphics[width=0.25\linewidth]{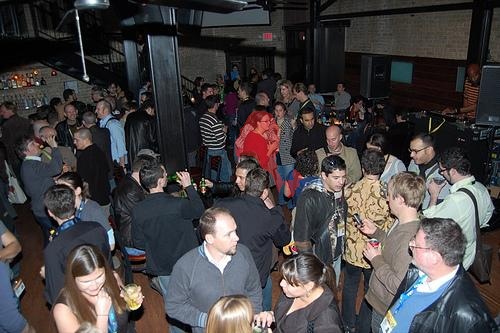} &
        \includegraphics[width=0.25\linewidth]{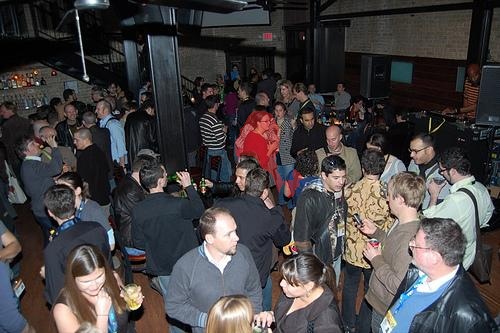}
    \end{tabular}

    \vspace{3mm}

    \parbox{0.98\linewidth}{\centering \small \textbf{Question 2:} Are the girl's eyes open or closed?}
    
    \vspace{1mm}
    
    \begin{tabular}{ccc}
        \includegraphics[width=0.25\linewidth]{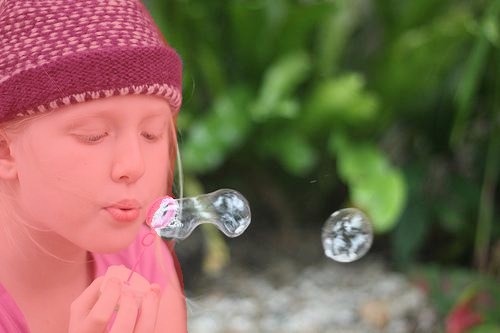} &
        \includegraphics[width=0.25\linewidth]{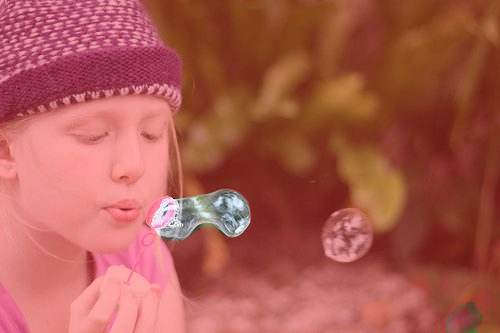} &
        \includegraphics[width=0.25\linewidth]{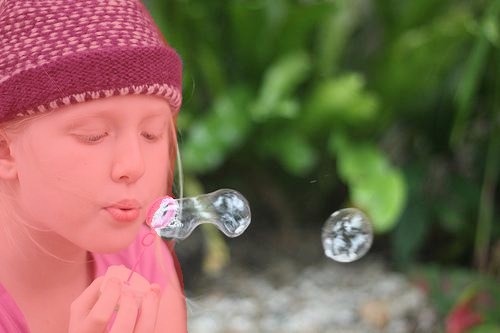}
    \end{tabular}

    \vspace{3mm}

    \caption{
    Comparison of masks produced independently by Qwen 3 and Qwen 2.5, together with the final consensus mask obtained by aggregating the judge-specific outputs. Across examples, the consensus mask preserves regions of agreement while filtering out judge-specific noise.
    }
    \label{fig:aggregation_comparison}
\end{figure*}

In figure \ref{fig:aggregation_comparison}, we compare the region labels produced by both judges against their consensus. While both judges often agree on the same portions of the image, it may be the case that one judge finds a region to be important and the other does not. We observe such cases for both judges; neither Qwen 3 32B nor Qwen 2.5 72B is consistently the more "overzealous" judge. When this situation arises, the use of intersection as an aggregation produces a narrower label.

\subsection{Comparison of Produced Labels}
\begin{figure*}[t]
    \centering
    \fbox{\parbox{0.95\textwidth}{\centering \small \textbf{Question 1:} What is the container being used to drink from?}}
    \vspace{0.6em}
    
    \begin{minipage}[t]{0.24\textwidth}
        \centering
        \includegraphics[width=\linewidth]{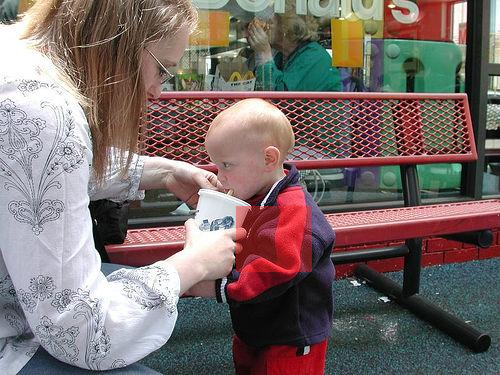}
        
        {\small \textbf{Qwen}}
    \end{minipage}
    \hfill
    \begin{minipage}[t]{0.24\textwidth}
        \centering
        \includegraphics[width=\linewidth]{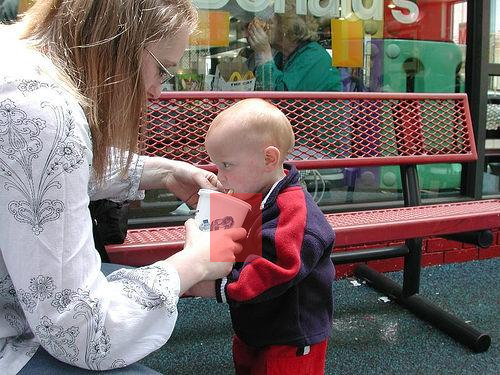}
        
        {\small \textbf{\textsc{MOLMO2}}}
    \end{minipage}
    \hfill
    \begin{minipage}[t]{0.24\textwidth}
        \centering
        \includegraphics[width=\linewidth]{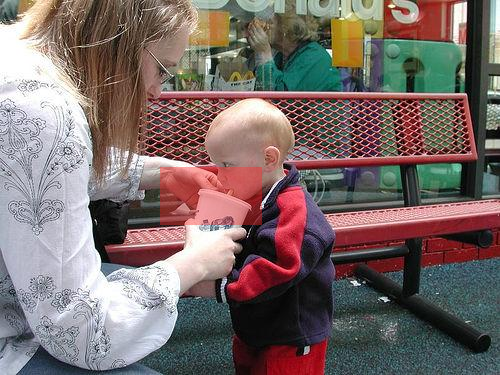}
        
        {\small \textbf{\textsc{VisCoT}}}
    \end{minipage}
    \hfill
    \begin{minipage}[t]{0.24\textwidth}
        \centering
        \includegraphics[width=\linewidth]{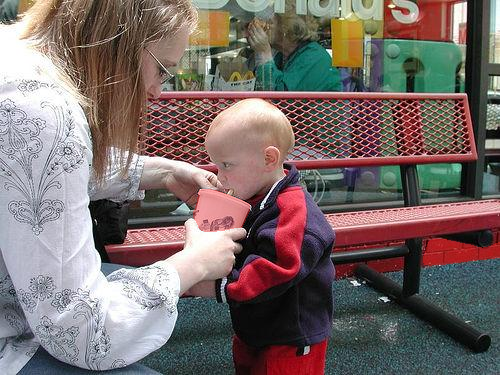}
        
        {\small \textbf{\method{}}}
    \end{minipage}
    \caption{
    Comparison of grounding labels produced by Qwen, \textsc{MOLMO2}, \textsc{VisCoT}, and \method{} across one example question-image pair.
    }
    \label{fig:label_comparison_one}
\end{figure*}

\begin{figure*}[t]
    \centering
    
    \fbox{\parbox{0.95\textwidth}{\centering \small \textbf{Question 2:} Does the truck in the image have ``V8/ Bomber'' written on its side?}}
    \vspace{0.6em}
    
    \begin{minipage}[t]{0.24\textwidth}
        \centering
        \includegraphics[width=\linewidth]{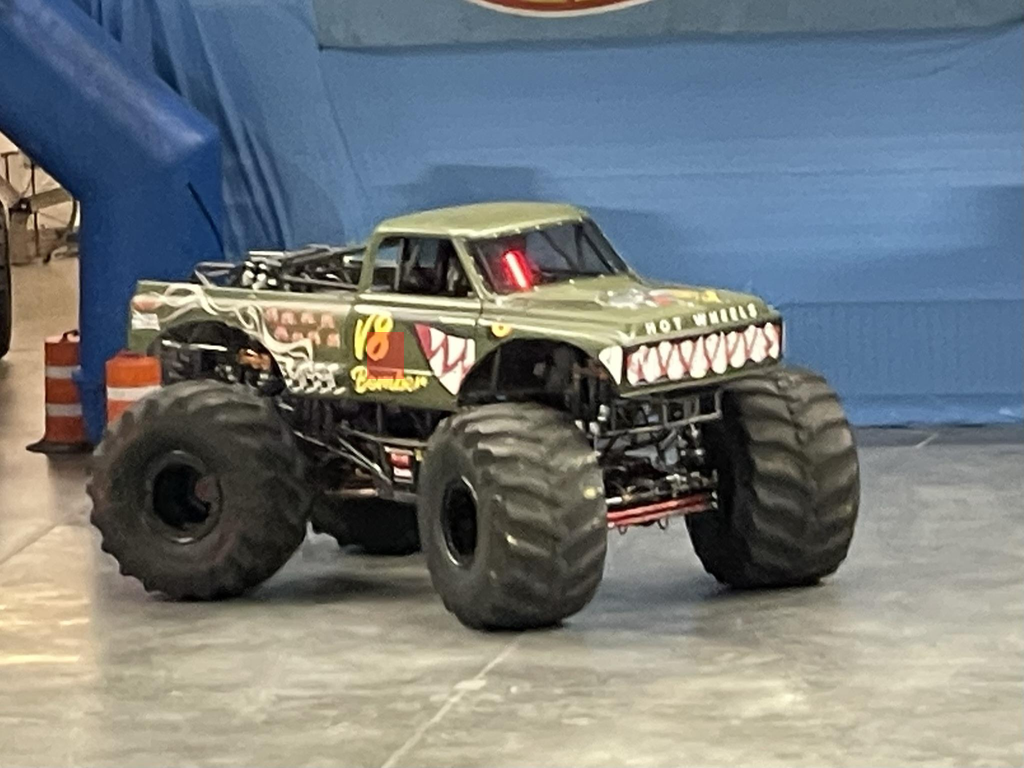}
        
        {\small \textbf{Qwen}}
    \end{minipage}
    \hfill
    \begin{minipage}[t]{0.24\textwidth}
        \centering
        \includegraphics[width=\linewidth]{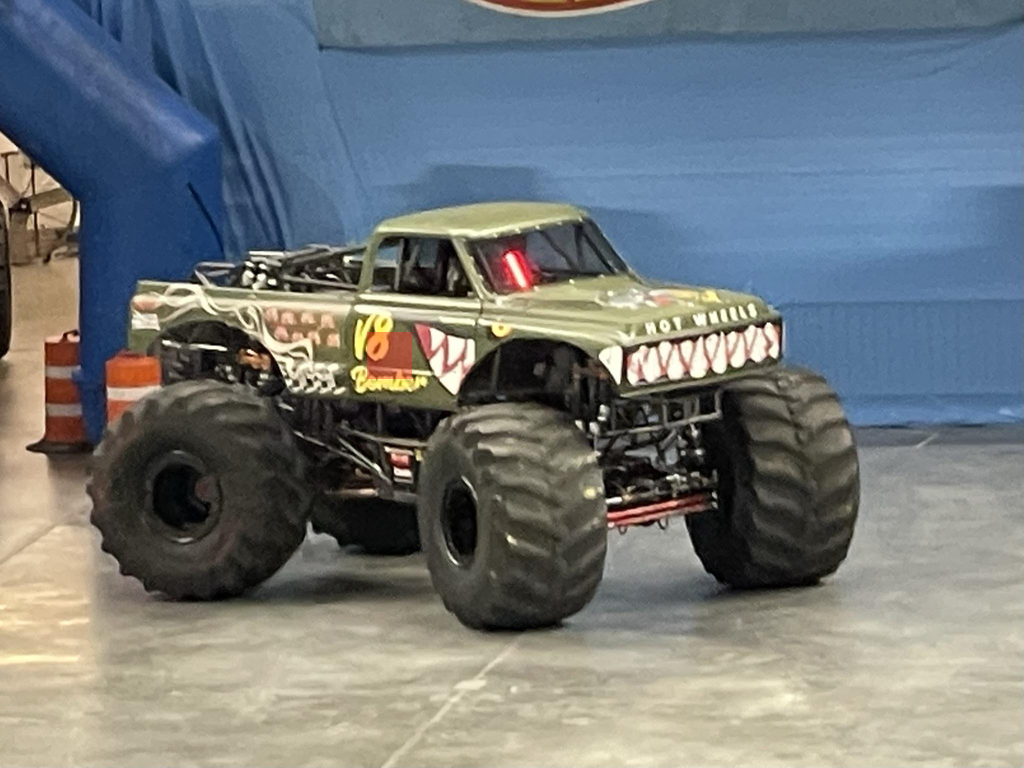}
        
        {\small \textbf{\textsc{MOLMO2}}}
    \end{minipage}
    \hfill
    \begin{minipage}[t]{0.24\textwidth}
        \centering
        \includegraphics[width=\linewidth]{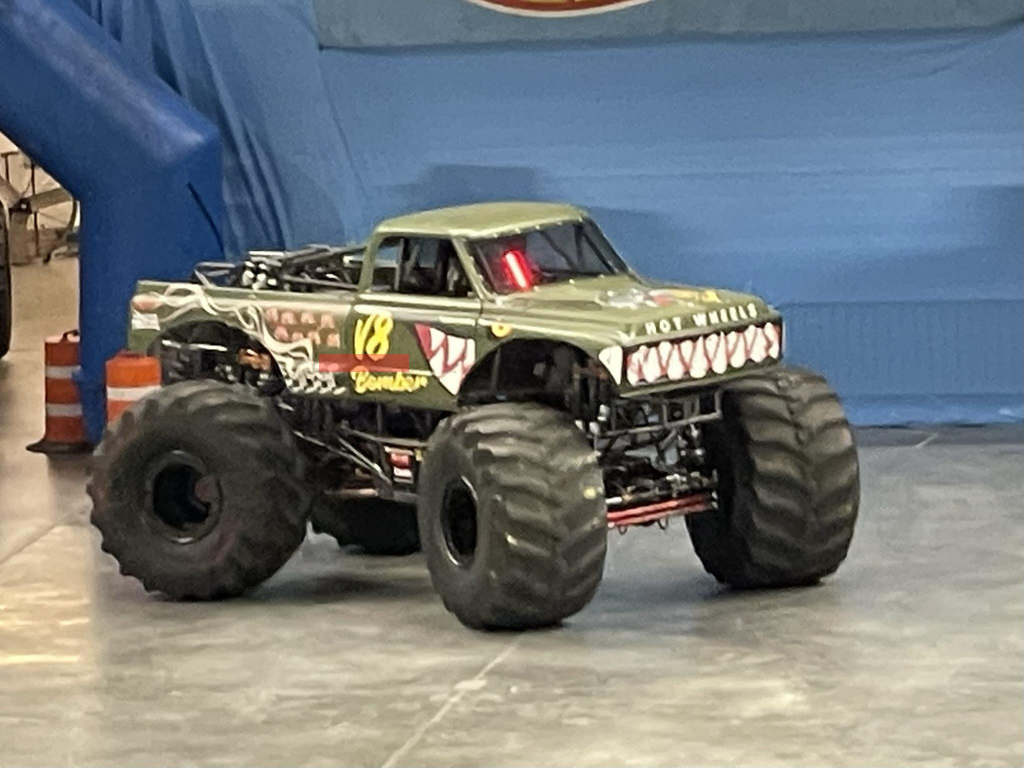}
        
        {\small \textbf{\textsc{VisCoT}}}
    \end{minipage}
    \hfill
    \begin{minipage}[t]{0.24\textwidth}
        \centering
        \includegraphics[width=\linewidth]{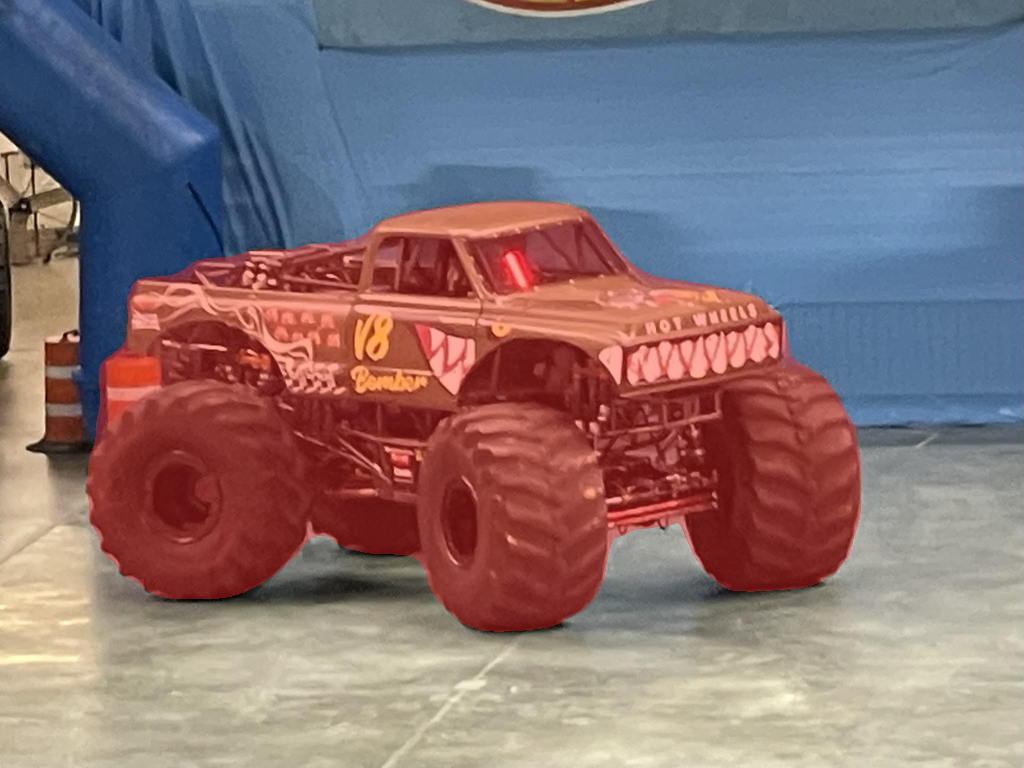}
        
        {\small \textbf{\method{}}}
    \end{minipage}

    \vspace{1em}
    
    \fbox{\parbox{0.95\textwidth}{\centering \small \textbf{Question 3:} Is there a dog involved in the activity?}}
    \vspace{0.6em}
    
    \begin{minipage}[t]{0.24\textwidth}
        \centering
        \includegraphics[width=\linewidth]{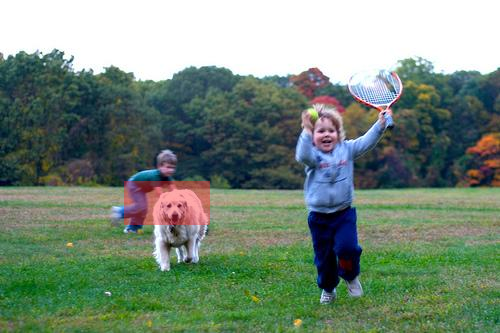}
        
        {\small \textbf{Qwen}}
    \end{minipage}
    \hfill
    \begin{minipage}[t]{0.24\textwidth}
        \centering
        \includegraphics[width=\linewidth]{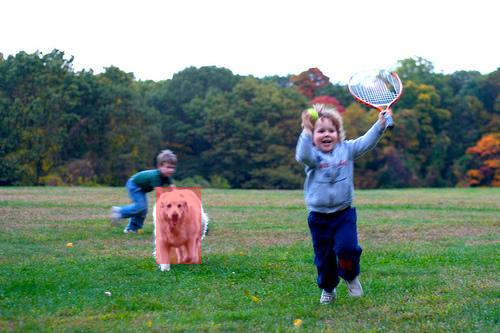}
        
        {\small \textbf{\textsc{MOLMO2}}}
    \end{minipage}
    \hfill
    \begin{minipage}[t]{0.24\textwidth}
        \centering
        \includegraphics[width=\linewidth]{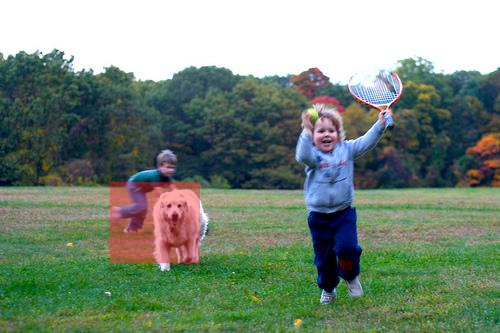}
        
        {\small \textbf{\textsc{VisCoT}}}
    \end{minipage}
    \hfill
    \begin{minipage}[t]{0.24\textwidth}
        \centering
        \includegraphics[width=\linewidth]{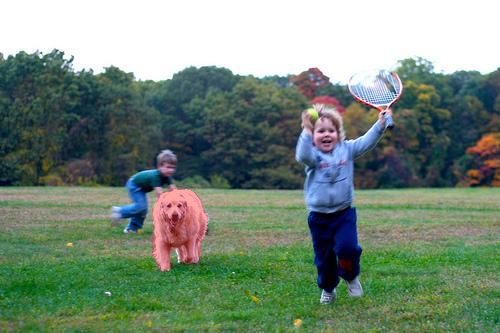}
        
        {\small \textbf{\method{}}}
    \end{minipage}

    \caption{
    Comparison of grounding labels produced by Qwen, \textsc{MOLMO2}, \textsc{VisCoT}, and \method{} across two example question-image pairs.
    }
    \label{fig:label_comparison_appndx}
\end{figure*}

Here, we provide a qualitative comparison of the four label sources. We see that \method{} provides a finer grained region label, coming from the use of a segmentation model for region proposal. In figures \ref{fig:label_comparison_one} and \ref{fig:label_comparison_appndx}, we see that the competing labels often cover a similar area, but are  shifted or do not cover the relevant region fully. \method{} instead produces masks that more tightly cover the answer-sensitive region identified by the counterfactual search. Figure \ref{fig:label_comparison_appndx} demonstrates the capability of all 4 sources to isolate a key region of the image, but that \method{} produces masks that more precisely cover the important objects in these examples.

\subsection{Question Dependent Labeling}
\begin{figure*}[t]
    \centering

    \begin{minipage}[t]{0.48\textwidth}
        \centering
        \fbox{\parbox{0.7\linewidth}{\centering \small \textbf{Question 1:} What is the container being used to drink from?}}
        \vspace{0.5em}
        
        \includegraphics[width=0.7\linewidth]{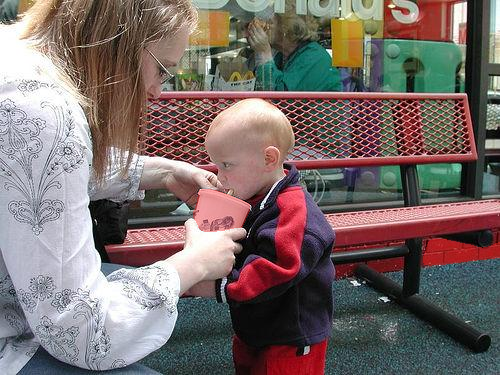}
        
    \end{minipage}
    \hfill
    \begin{minipage}[t]{0.48\textwidth}
        \centering
        \fbox{\parbox{0.7\linewidth}{\centering \small \textbf{Question 2:} Who is assisting the child with the drink?}}
        \vspace{0.5em}
        
        \includegraphics[width=0.7\linewidth]{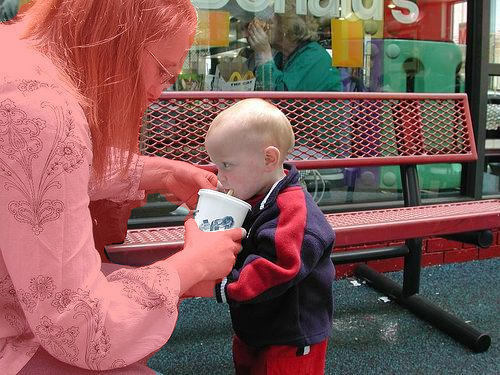}
        
    
    \end{minipage}

    \caption{
    Model-causal masks found for the same underlying image under different questions, illustrating how the discovered evidence changes with the query.
    }
    \label{fig:same_image_different_questions}
\end{figure*}

Figure \ref{fig:same_image_different_questions} shows that \method{} produces question-dependent model-causal masks. For the same image, the discovered regions change across questions, indicating that the method is not simply selecting salient objects, but identifying regions whose perturbation affects the model's answer for the current query.

\subsection{Figures}

\begin{figure}[h]
    \centering

    \fbox{\parbox{0.95\linewidth}{\centering \small \textbf{Question 1:} From the angle of the picture, is there a shadow of a car on the left side of the BMW? (Options: Yes, No)}}
    \vspace{0.5em}

    \includegraphics[width=0.4\linewidth]{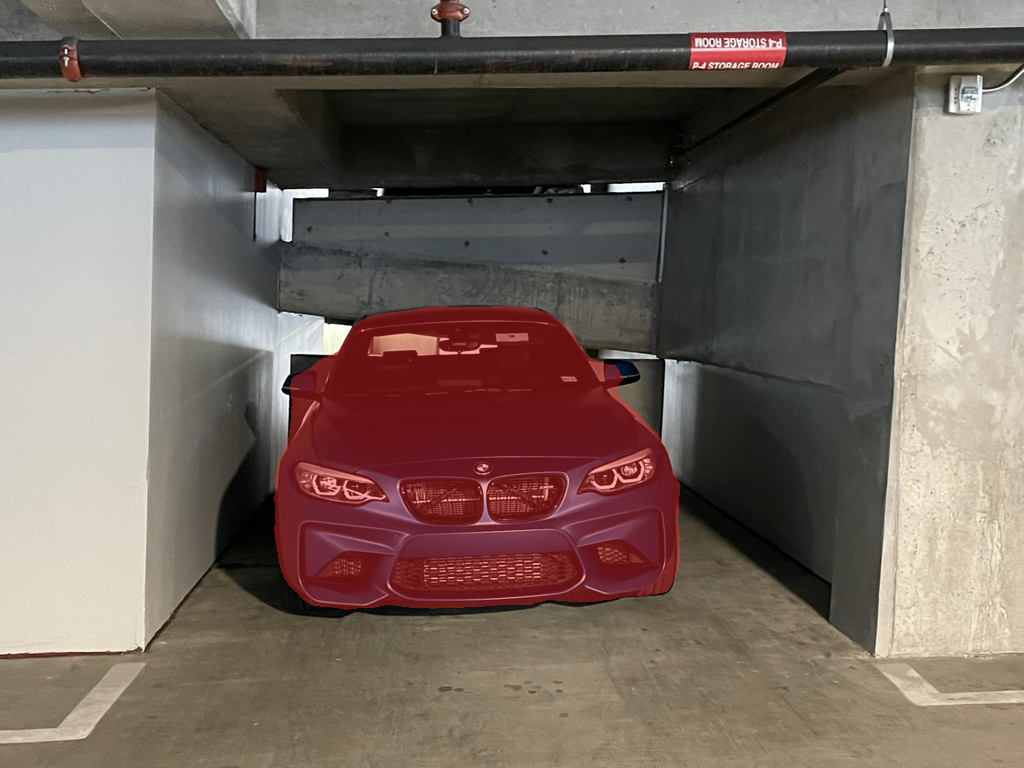}

    \vspace{1em}

    \fbox{\parbox{0.95\linewidth}{\centering \small \textbf{Question 2:} How many people are in the scene? (Options: 1, more than 3)}}
    \vspace{0.5em}

    \includegraphics[width=0.25\linewidth]{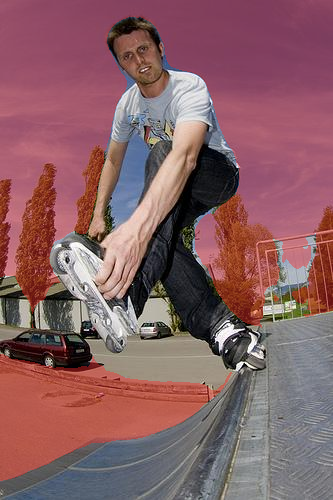}

    \caption{
    Example failure cases for \method{}. In the first example, the predicted grounding highlights the vehicle, but the answer depends on the shadow it casts. In the second example, for a counting problem, \method{} identifies regions where adding new background items can change the answer, even though the correct evidence should be the people already present in the scene.
    }
    \label{fig:failure_cases}
\end{figure}

\end{document}